\documentclass[sigconf]{acmart}

\usepackage{booktabs} 
\usepackage{hyperref}       
\usepackage{url}            
\usepackage{booktabs}       
\usepackage{amsfonts}       
\usepackage{nicefrac}       
\usepackage{microtype}      
\usepackage{xcolor}         
\usepackage{diagbox} 
\usepackage{multirow} 
\usepackage{colortbl} 
\usepackage{pifont} 
\usepackage{makecell} 
\usepackage{enumitem}
\usepackage{tabularx}
\usepackage{array}
\usepackage{msc}
\usepackage{subcaption}

\usepackage[ruled,linesnumbered]{algorithm2e} 

\SetAlFnt{\small}
\SetAlCapFnt{\small}
\SetAlCapNameFnt{\small}
\SetAlCapHSkip{0pt}

\setcopyright{acmcopyright}

\setcopyright{rightsretained}
\copyrightyear{2025}
\acmYear{2025}
\acmConference[SA Posters '25]{SIGGRAPH Asia 2025 Posters}{December 15--18, 2025}{Hong Kong, Hong Kong}
\acmBooktitle{SIGGRAPH Asia 2025 Posters (SA Posters '25), December 15--18, 2025}
\acmDOI{10.1145/3757374.3771455}
\acmISBN{979-8-4007-2134-2/2025/12}

\begin{document}
\title{AniME: Adaptive Multi-Agent Planning for Long Animation Generation}

\author{Lisai Zhang, Baohan Xu, Siqian Yang, Mingyu Yin, Jing Liu, Chao Xu, Siqi Wang, Yidi Wu, Yuxin Hong, Zihao Zhang, Yanzhang Liang, Yudong Jiang}
\affiliation{%
 \institution{Bilibili. Inc}
 \country{China}
}
\authornote{Corresponding author to nebuladream@gmail.com}


\renewcommand\shortauthors{Zhang. et al}

\begin{abstract}
We present \textbf{AniME}, a director-oriented multi-agent system for automated long-form anime production, covering the full workflow from a story to the final video. The director agent keeps a global memory for the whole workflow, and coordinates several downstream specialized agents. By integrating customized Model Context Protocol (MCP) with downstream model instruction, the specialized agent adaptively selects control conditions for diverse sub-tasks. AniME produces cinematic animation with consistent characters and synchronized audio–visual elements, offering a scalable solution for AI-driven anime creation.

\end{abstract}


%
%


%
%


\maketitle

\begin{figure}
    \centering
    \includegraphics[width=1\linewidth]{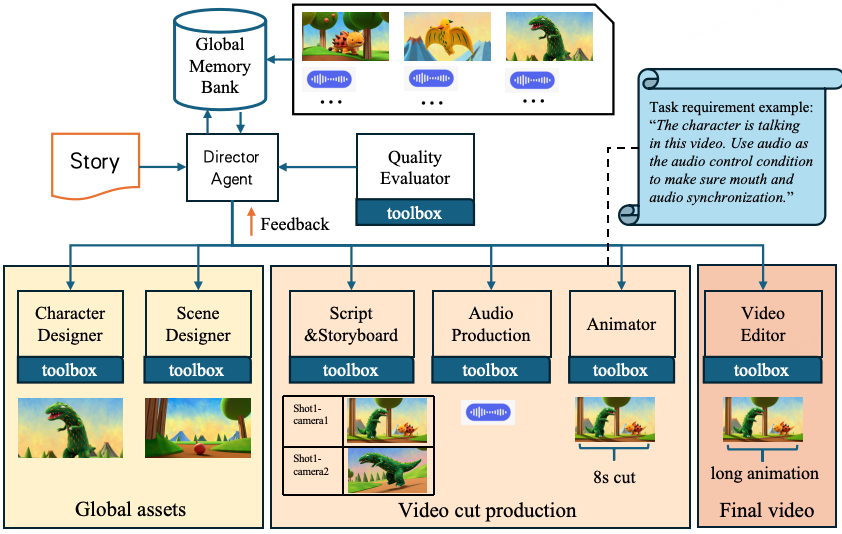}
    \caption{Brief Architecture of AniME}
    \label{fig:placeholder}
    \vspace{-5pt}
\end{figure}

\section{Introduction}

Anime production is a complex, labor-intensive process evolves multiple stages such as scriptwriting, storyboarding, character and scene design, animation, voice acting, and final editing. Traditional workflows require extensive manual expertise and close collaboration across diverse teams, resulting in high costs and long production cycles. Recent advances in generative AI, such as AniSora~\cite{tool-jiang2024anisora} for animation generation, have demonstrated impressive capabilities in specific tasks~\cite{tool-du2025vall,tool-majumder2024tango}. However, these methods each exhibit distinct strengths and weaknesses in particular domains, leading to challenges in maintaining consistency~\cite{llm-yang2024mastering,agent-shi2025animaker} and achieving fine-grained controllability in agent-driven video generation~\cite{agent-li2024anim,agent-wu2025automated,agent-xia2025storywriter}. As a result, developing a fully automated system for long-form anime generation remains an open challenge, particularly in selecting appropriate control conditions and ensuring cross-stage content consistency.

In this work, we present \emph{AniME}, a novel director-oriented multi-agent framework that integrates specialized agents equipped with tailored Model Context Protocol (MCP)~\cite{tool-mcp} toolsets. 
The framework employs centralized planning and quality control to coordinate task scheduling and ensure content consistency across stages. 
It enables coherent long-form video generation through a combination of rich, controllable generative models and iterative feedback workflows. 
Furthermore, we customize the MCP protocol of downstream generative tools to explicitly annotate each tool’s domain expertise and limitations, allowing agents to select appropriate control condition models for specific sub-tasks.


\begin{table*}[h]
\centering
\small
\caption{Summary of Specialized Agents and their MCP Toolsets}
\vspace{-5pt}
\label{tab:agents}
\begin{tabularx}{\linewidth}{ 
  >{\raggedright\arraybackslash}p{2.5cm} 
  >{\raggedright\arraybackslash}p{4cm} 
  >{\raggedright\arraybackslash}p{4cm} 
  >{\raggedright\arraybackslash}X 
}
\toprule
\textbf{Agent} & \textbf{Inputs} & \textbf{Outputs} & \textbf{Tools \& Methods} \\
\hline

Character Designer & 
visual style, character description & 
character multi-view image  & 
Text-to-image + refinement; multi-view synthesis \\

\hline

Scene Designer & 
visual style, Scene description& 
Background images, layered assets & 
Depth-guided image generator; layout-guided image generator; relighting model\\

\hline

Script \& Storyboard & 
Shot description text & 
Timeline, shot prompt, keyframes, camera motion & 
LLM-based segmentation and tagging; camera planner; layout planner; reference image generation \\

\hline

Animator & 
Keyframes/camera paths/rigs/poses/audio & 
Frame sequences  & 
Keyframe/audio/pose/camera conditioned video generation model  \\

\hline

Audio Production & 
Dialogue with emotion labels, scene tags & 
Phoneme-aligned audio & 
speaker conditioned TTS; text/video to music model; audio mixer programs \\

\hline

Video Editor & 
Frame sequences, audio stems, editorial instructions & 
Final encoded video & 
Transition effects; color pipeline; FFmpeg multi-pass encoding \\

\hline

Quality Evaluator & 
Generated frames and assets & 
text-to-video similarity & 
text-to-image similarity; identity verification; AV sync checks; VLM narrative evaluator \\

\bottomrule
\end{tabularx}
\vspace{-5pt}
\end{table*}

\section{Method}
\label{sec:method}
AniME decomposes the story-to-video task into hierarchical stages coordinated by a centralized \textbf{Director Agent} and executed by a set of \textbf{Specialized Agents}. Each agent $A_i$ has a well-defined input type $\mathcal{I}_i$, output type $\mathcal{O}_i$, and a local MCP toolbox $\mathcal{T}_i$. 

\subsection{Director Agent}

The Director Agent serves as the central controller of the AniME framework, managing both the global workflow and quality assurance. It decomposes the input story into a task workflow, allocates subtasks to agents, checks the quality of each specialized agent, and maintains a global Asset Memory Bank.

\begin{algorithm}[h]
\caption{AniME: Multi-Agent Workflow}
\label{alg:animegen_workflow}

\KwIn{Story script \(\mathcal{R}\)}
\KwOut{Final video \(\mathcal{V}\)}

Initialize Director with \(\mathcal{R}\), derive shots \(\mathcal{H}\) and styles \(\mathbf{s}_v, \mathbf{s}_a\)\;
Generate task list \(T\) and task dependencies graph \(\mathcal{W} = (N, E)\)\;

\ForEach{task \(t \in T\) in topological order of $\mathcal{W}$}{
  Assign \(t\) to agent \(A_i\)\;
  Agent $A_i$ select proper model from MCP toolset $\mathcal{T}_i$ according to task requirement;
  
  Agent $A_i$  producing output $\mathcal{O}_i(t)$\ and return to Director; 
  
  \If{Evaluated quality of \(\mathcal{O}_i(t)\)\ low}{
    Request revision
  }
  Store \(\mathcal{O}_i(t)\) in Asset Memory\;
}

Director performs final editing via Video Editor Agent\;

\Return \(\mathcal{V}\)\;

\end{algorithm}

\subsubsection{Workflow of the Director}
Given a long-form story $\mathcal{R}$, the Director initiates a hierarchical breakdown into scenes and shots through a segmentation process, and decides the style for visual $\mathbf{s}_v$ and $\mathbf{s}_v$ acoustic setting. Then it uses chain-of-thought prompts to generate an initial task list $T$. 

The Director maintains a workflow graph $\mathcal{W} = (N, E)$, where each node $n \in N$ corresponds to a production task (e.g., ``generate character pose for shot 3'') and edges $E$ encode explicit dependencies (e.g., \textit{character design} $\rightarrow$ \textit{storyboard} $\rightarrow$ \textit{animation}). 

\begin{table}[h]
\centering
\caption{ Names and key fields for the Asset Memory Table}

\small
\begin{tabular}{l l}
\hline
\textbf{Table Name} & \textbf{Keys} \\
\hline
shot & id, description \\
scene & id, prompt, view\_3d \\
character & id, prompt, demo\_voice, voice\_prompt, 3d\_view \\
 style & id, visual\_style, acoustic\_style \\
storyboard & id,  prompt, image\_path \\
video & id,  prompt, video\_path, shot\_id, music\_id \\
music & id, character\_id,  prompt, music\_path \\
\hline
\end{tabular}
\label{tab:asset}
\vspace{-5pt}
\end{table}

\subsubsection{Asset Memory Management}
The Asset Memory Management module stores and organizes key creative assets across the production pipeline, ensuring consistency and reusability. The assets are shown in Table ~\ref{tab:asset}. The management is implemented using queryable database tables, enabling efficient indexing, retrieval, and update operations. 



\subsection{Specialized Agents and MCP Tools Selection}

AniME employs a set of specialized agents designed for a key production stage with dedicated MCP toolsets. The Table~\ref{tab:agents} demonstrates a detailed overview of each agent's inputs, outputs, and tools. Specifically, we describe the advantages and disadvantages of each tool in the MCP protocol, enabling the agent to decide on the task requirements. 

The \textit{Script \& Storyboard} transforms narrative scripts into structured shot descriptors and visual keyframes with camera plans. The \textit{Character Designer} generates canonical character assets, including multi-view portraits and identity embeddings. The \textit{Scene Designer} creates backgrounds and environmental assets consistent with scene perspective and style. The \textit{Animator} synthesizes animated frame sequences and performs precise lip-syncing, maintaining character identity and style coherence. The \textit{Audio Production} handles expressive speech synthesis, ambient sound generation, music composition, and audio mixing. The \textit{Video Editor} integrates visual and audio components, applies color grading and transitions, and produces the final encoded video. Finally, the \textit{Quality Evaluator} automates multi-modal verification of visual and narrative coherence, triggering targeted revisions when necessary. 

Meanwhile, animation creators can interact with the agent at any stage, guiding the agent to produce animations that better align with the creator’s artistic style.

\section{Conclusion}
We presented \emph{AniME}, a director-oriented multi-agent framework for automated long-form anime production. By integrating specialized agents equipped with customized MCP toolsets, the agents adaptively decide the optimistic available models and achieved consistent generation from a story to the final video.

\bibliographystyle{ACM-Reference-Format}
\bibliography{reference}


\begin{thebibliography}{9}


\ifx \showCODEN    \undefined \def \showCODEN     #1{\unskip}     \fi
\ifx \showDOI      \undefined \def \showDOI       #1{#1}\fi
\ifx \showISBNx    \undefined \def \showISBNx     #1{\unskip}     \fi
\ifx \showISBNxiii \undefined \def \showISBNxiii  #1{\unskip}     \fi
\ifx \showISSN     \undefined \def \showISSN      #1{\unskip}     \fi
\ifx \showLCCN     \undefined \def \showLCCN      #1{\unskip}     \fi
\ifx \shownote     \undefined \def \shownote      #1{#1}          \fi
\ifx \showarticletitle \undefined \def \showarticletitle #1{#1}   \fi
\ifx \showURL      \undefined \def \showURL       {\relax}        \fi
\providecommand\bibfield[2]{#2}
\providecommand\bibinfo[2]{#2}
\providecommand\natexlab[1]{#1}
\providecommand\showeprint[2][]{arXiv:#2}

\bibitem[Anthropic(2024)]%
        {tool-mcp}
\bibfield{author}{\bibinfo{person}{Anthropic}.} \bibinfo{year}{2024}\natexlab{}.
\newblock \bibinfo{title}{Introducing the Model Context Protocol}.
\newblock
\newblock
\urldef\tempurl%
\url{http://www.anthropic.com/news/model-context-protocol}
\showURL{%
Retrieved Aug 18, 2024 from \tempurl}


\bibitem[Du et~al\mbox{.}(2025)]%
        {tool-du2025vall}
\bibfield{author}{\bibinfo{person}{Chenpeng Du}, \bibinfo{person}{Yiwei Guo}, \bibinfo{person}{Hankun Wang}, {et~al\mbox{.}}} \bibinfo{year}{2025}\natexlab{}.
\newblock \showarticletitle{Vall-t: Decoder-only generative transducer for robust and decoding-controllable text-to-speech}. In \bibinfo{booktitle}{\emph{ICASSP}}.
\newblock


\bibitem[Jiang et~al\mbox{.}(2024)]%
        {tool-jiang2024anisora}
\bibfield{author}{\bibinfo{person}{Yudong Jiang}, \bibinfo{person}{Baohan Xu}, \bibinfo{person}{Siqian Yang}, {et~al\mbox{.}}} \bibinfo{year}{2024}\natexlab{}.
\newblock \showarticletitle{Anisora: Exploring the frontiers of animation video generation in the sora era}.
\newblock \bibinfo{journal}{\emph{arXiv:2412.10255}} (\bibinfo{year}{2024}).
\newblock


\bibitem[Li et~al\mbox{.}(2024)]%
        {agent-li2024anim}
\bibfield{author}{\bibinfo{person}{Yunxin Li}, \bibinfo{person}{Haoyuan Shi}, \bibinfo{person}{Baotian Hu}, {et~al\mbox{.}}} \bibinfo{year}{2024}\natexlab{}.
\newblock \showarticletitle{Anim-director: A large multimodal model powered agent for controllable animation video generation}. In \bibinfo{booktitle}{\emph{SIGGRAPH Asia}}.
\newblock


\bibitem[Majumder et~al\mbox{.}(2024)]%
        {tool-majumder2024tango}
\bibfield{author}{\bibinfo{person}{Navonil Majumder}, \bibinfo{person}{Chia-Yu Hung}, \bibinfo{person}{Deepanway Ghosal}, {et~al\mbox{.}}} \bibinfo{year}{2024}\natexlab{}.
\newblock \showarticletitle{Tango 2: Aligning diffusion-based text-to-audio generations through direct preference optimization}. In \bibinfo{booktitle}{\emph{ACM MM}}.
\newblock


\bibitem[Shi et~al\mbox{.}(2025)]%
        {agent-shi2025animaker}
\bibfield{author}{\bibinfo{person}{Haoyuan Shi}, \bibinfo{person}{Yunxin Li}, \bibinfo{person}{Xinyu Chen}, {et~al\mbox{.}}} \bibinfo{year}{2025}\natexlab{}.
\newblock \showarticletitle{AniMaker: Automated Multi-Agent Animated Storytelling with MCTS-Driven Clip Generation}.
\newblock \bibinfo{journal}{\emph{arXiv:2506.10540}} (\bibinfo{year}{2025}).
\newblock


\bibitem[Wu et~al\mbox{.}(2025)]%
        {agent-wu2025automated}
\bibfield{author}{\bibinfo{person}{Weijia Wu}, \bibinfo{person}{Zeyu Zhu}, {and} \bibinfo{person}{Mike~Zheng Shou}.} \bibinfo{year}{2025}\natexlab{}.
\newblock \showarticletitle{Automated movie generation via multi-agent cot planning}.
\newblock \bibinfo{journal}{\emph{arXiv:2503.07314}} (\bibinfo{year}{2025}).
\newblock


\bibitem[Xia et~al\mbox{.}(2025)]%
        {agent-xia2025storywriter}
\bibfield{author}{\bibinfo{person}{Haotian Xia}, \bibinfo{person}{Hao Peng}, \bibinfo{person}{Yunjia Qi}, {et~al\mbox{.}}} \bibinfo{year}{2025}\natexlab{}.
\newblock \showarticletitle{StoryWriter: A Multi-Agent Framework for Long Story Generation}.
\newblock \bibinfo{journal}{\emph{arXiv:2506.16445}} (\bibinfo{year}{2025}).
\newblock


\bibitem[Yang et~al\mbox{.}(2024)]%
        {llm-yang2024mastering}
\bibfield{author}{\bibinfo{person}{Ling Yang}, \bibinfo{person}{Zhaochen Yu}, \bibinfo{person}{Chenlin Meng}, {et~al\mbox{.}}} \bibinfo{year}{2024}\natexlab{}.
\newblock \showarticletitle{Mastering Text-to-Image Diffusion: Recaptioning, Planning, and Generating with Multimodal LLMs}. In \bibinfo{booktitle}{\emph{ICML}}.
\newblock


\end{thebibliography}

\appendix
\section{Method Details}
\label{sec:method2}

AniME decomposes the story-to-video task into hierarchical stages coordinated by a centralized \textbf{Director Agent} and executed by a set of \textbf{Specialized Agents}. Each agent $A_i$ has a well-defined input type $\mathcal{I}_i$, output type $\mathcal{O}_i$, and a local MCP toolbox $\mathcal{T}_i$. Communication between agents is performed via structured JSON messages. In this section, we detail the overall framework, the role of the Director, asset memory management, inter-agent communication, and the specialized agents that together enable long-form animation production.

\subsection{Director Agent}
The Director Agent serves as the central controller of AniME. It manages the global workflow, decomposes the input story into subtasks, allocates them to specialized agents, evaluates quality, and maintains the global Asset Memory Bank. Unlike traditional pipeline orchestration, the Director employs hierarchical planning with explicit dependency tracking to preserve both narrative coherence and visual consistency.

\subsubsection{Workflow of the Director}
Given a story script $\mathcal{R}$, the Director first segments it into scenes and shots. The segmentation uses semantic similarity across narrative units, emotional state transitions, and temporal markers (e.g., dialogue breaks, action beats). Each scene inherits a global style vector for visuals $\mathbf{s}_v$ and acoustics $\mathbf{s}_a$, determined through a style analysis model.

The Director then generates a task list $T$ and constructs a workflow graph $\mathcal{W} = (N, E)$, where nodes represent subtasks (e.g. ``render background for shot 5'') and edges represent dependencies (e.g., character design precedes animation). Tasks are executed in topological order, with failed or low-quality outputs triggering localized revisions without re-running the full pipeline.

\subsubsection{Asset Memory Management}
The Asset Memory Bank stores canonical creative assets, ensuring consistency and reusability across the pipeline. Each table is queryable, versioned, and embedding-augmented, enabling retrieval by semantic similarity. For example, the Character table stores not only reference images but also identity embeddings, ensuring that the Animator preserves visual fidelity. Table~\ref{tab:asset} lists representative tables used in the memory.

To prevent drift, the Director enforces single-writer semantics for canonical fields (e.g., only the Character Designer can update identity vectors), while enabling multi-reader access for dependent agents. Version control is integrated, allowing rollback or branching when user-guided revisions are made.

\subsection{Inter-Agent Communication Protocol}
All agents communicate using structured JSON messages. 
Each request message contains task identifiers, task description, assets, and metadata. A typical task request message is as follows.
\begin{verbatim}
{
"id": "shot_03_v1",
"type": "storyboard",
"task": {
    "prompt": "split the given shot into segments,  
    plan camera angles, and generate the keyframes. ",
"requirement": "style: should be 3d animation"},
"assets": { "style": [...], "identity": [...] },
"meta": { "version": 1, "producer": "StoryboardAgent" }
}
\end{verbatim}
This structure supports message versioning, multimodal embedding sharing, and agent provenance tracking. Agents can query the Asset Memory with message IDs or embedding similarity, enabling flexible yet consistent coordination.

The response messages contain the generated outputs, including asset IDs, visual aids, and metadata. For example:
\begin{verbatim}
{
"id": "shot_03_v1",
"type": "storyboard",
"status": "success",
"outputs": {
    "keyframes": ["kf1.png", "kf2.png"],
    "camera_plan": {
        "angles": [30, 45],
        "transitions": ["fade", "cut"]
    }
},
"meta": { "version": 1, "producer": "StoryboardAgent" }
}
\end{verbatim}
This response structure allows agents to validate outputs, track dependencies, and maintain a coherent workflow.

\subsection{Specialized Agents and MCP Toolsets}

\begin{table*}[t]
\centering
\small
\begin{tabular}{p{2cm}p{4cm}p{5cm}p{4cm}}
\hline
\textbf{Tool Type} & \textbf{Functionality} & \textbf{Pros} & \textbf{Cons} \\
\hline

\textbf{Text-to-Image} &
Generate images directly from text prompts, useful for establishing shots or background scenes. &
\begin{itemize}[leftmargin=*]
  \item Simple interface
  \item Good for empty-scene (establishing) shots
\end{itemize} &
\begin{itemize}[leftmargin=*]
  \item Hard to control layout
  \item Limited consistency across shots
\end{itemize} \\
\hline

\textbf{Reference Image Generation} &
Generate images conditioned on reference characters to ensure identity consistency across shots. &
\begin{itemize}[leftmargin=*]
  \item Ensures character consistency
  \item Good for dialogue-focused panels
\end{itemize} &
\begin{itemize}[leftmargin=*]
  \item Cannot generate empty establishing shots
\end{itemize} \\
\hline

\textbf{Layout-Guided Image Generation} &
Generate images following a predefined layout or sketch, suitable for precise camera and composition control. &
\begin{itemize}[leftmargin=*]
  \item Strong spatial control
  \item Useful for complex multi-character panels
\end{itemize} &
\begin{itemize}[leftmargin=*]
  \item Higher complexity
  \item Requires layout design overhead
\end{itemize} \\
\hline

\end{tabular}
\caption{Storyboard-related MCP tools example. Different generation types (text-to-image, reference image generation, layout-guided image generation) are adaptively select by the Script and Storyboard Agent according to the generation task requirements.}
\label{tab:storyboard-tools}
\end{table*}

AniME employs a set of specialized agents, each aligned to a creative stage. Each agent leverages a dedicated MCP toolset, allowing adaptive model selection based on task requirements. Below, we expand on their detailed designs.

\subsubsection{Script and Storyboard Agent}
The Script and Storyboard Agent transforms narrative text into a temporal sequence of shots. It performs three key steps: (1) \emph{segmentation} of text into scenes and shots using discourse parsing and emotion tagging; (2) \emph{camera planning}, including shot type, trajectory, and transitions; and (3) \emph{reference retrieval}, where keyframes are generated using retrieval-augmented text-to-image models. The outputs are structured JSON shot descriptors with the corresponding visual aids.

\textbf{Tool selection.} This agent adaptively selects among multiple tools to generate storyboards: 
\emph{text-to-image} for empty establishing shots (simple, fast, but limited layout control), 
\emph{reference image generation} for maintaining character identity consistency (strong on character fidelity but cannot generate empty scenes), and 
\emph{layout-guided generation} for precise multi-character or object-dense panels (accurate composition but higher computational cost). LLM-based segmentation, camera planning, and layout planning modules orchestrate the shot-level structure.

\subsubsection{Character Designer}
The Character Designer creates canonical character assets. Starting from textual descriptions and style vectors, it generates multi-view portraits via diffusion-based text-to-image models. Identity consistency is enforced by CLIP-based scoring against stored embeddings. Multi-view synthesis further ensures that side and back profiles match the canonical frontal view. Failure detection is applied to reject outputs with drifted identity or inconsistent clothing details.

\textbf{Tool selection.} The agent mainly leverages \emph{reference image generation} to ensure consistent character identity. \emph{Text-to-image} is used for initial exploration or design iteration, while \emph{multi-view synthesis} and refinement modules enforce multi-angle consistency and correct visual drift.

\subsubsection{Scene Designer}
The Scene Designer generates environments and backgrounds. It uses depth-guided diffusion for spatial coherence, layout-guided generation for furniture and object placement, and relighting models for temporal consistency across shots. Outputs include layered assets that can be reused across multiple scenes.

\textbf{Tool selection.} Scene Designer prioritizes \emph{layout-guided generation} for precise object placement, \emph{depth-guided image generation} for spatially coherent scenes, and relighting models to maintain temporal lighting consistency. \emph{Text-to-image} can optionally generate broad establishing shots but provides less control over composition.

\subsubsection{Animator}
The Animator synthesizes motion sequences from keyframes, poses, and camera trajectories. It integrates keyframe-conditioned video diffusion with audio-driven lip synchronization. To maintain temporal coherence, optical flow guidance and motion interpolation are applied. The Animator also aligns phoneme embeddings from dialogue audio with mouth movements, ensuring expressive yet identity-preserving animation.

\textbf{Tool selection.} Animator mainly uses keyframe/audio/pose/camera-conditioned video generation models. In revision cycles, it may request additional layout-guided or reference image outputs from upstream agents to repair missing or inconsistent frames, but it generally does not directly invoke text-to-image tools.

\subsubsection{Audio Production Agent}
The Audio Production Agent manages dialogue, sound effects, and music. It employs speaker-conditioned TTS to generate character-specific voices, enriched with emotion embeddings. Background music is composed via text-to-music generation and synchronized with scene dynamics. Foley and ambient effects are retrieved from external databases or synthesized directly. An audio mixer then balances speech, music, and effects.

\textbf{Tool selection.} This agent selects among \emph{speaker-conditioned TTS}, \emph{text-to-music generation}, and \emph{audio mixer programs}. Each tool complements others: TTS ensures identity-preserving dialogue, text-to-music adapts to scene dynamics, and the mixer balances all audio tracks for coherent output.

\subsubsection{Video Editor Agent}
The Video Editor integrates all assets into a coherent final video. Automated editing tools suggest cuts and transitions, while a color pipeline ensures visual consistency. Multi-pass encoding via FFmpeg produces the final distribution-ready video. Human-in-the-loop revision is supported at this stage for fine-grained adjustments.

\textbf{Tool selection.} Video Editor primarily relies on \emph{transition effects}, \emph{color pipelines}, and \emph{FFmpeg multi-pass encoding}. While it does not generate new content, it consumes outputs from other agents to produce a professional-quality video and apply global visual adjustments.

\subsubsection{Quality Evaluator Agent}
The Quality Evaluator performs multimodal validation. It checks text-to-video similarity to ensure semantic alignment, identity verification to prevent drift, and audio-visual synchronization. Additionally, a vision-language model (VLM) is used for narrative consistency evaluation, ensuring that the generated sequence adheres to the intended storyline. When failures are detected, targeted revision requests are sent back to the responsible agent.

\textbf{Tool selection.} The evaluator uses \emph{text-to-video similarity scoring}, \emph{identity verification}, \emph{AV sync checks}, and \emph{VLM-based narrative evaluation}. Based on detected errors, it may recommend specific tool changes, such as switching storyboard generation from text-to-image to layout-guided for better spatial composition.

\subsubsection{MCP Tool Selection}

\begin{figure*}[t]
    \centering
    \begin{subfigure}[b]{0.32\textwidth}
        \includegraphics[width=\textwidth]{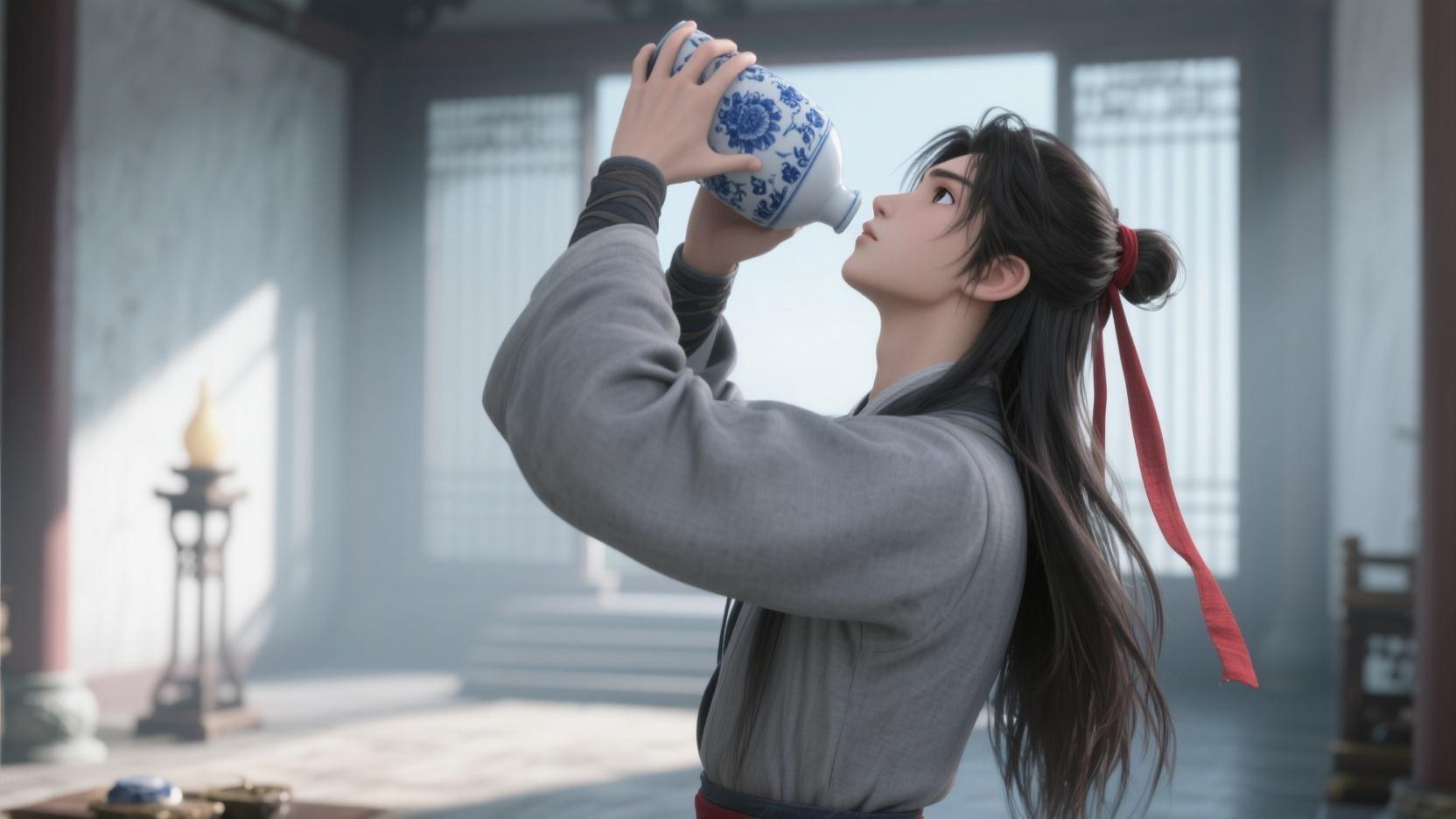}
        \caption{sceneYX01\_shot\_01}
        \label{fig:sceneYX01_shot_01}
    \end{subfigure}
    \hfill
    \begin{subfigure}[b]{0.32\textwidth}
        \includegraphics[width=\textwidth]{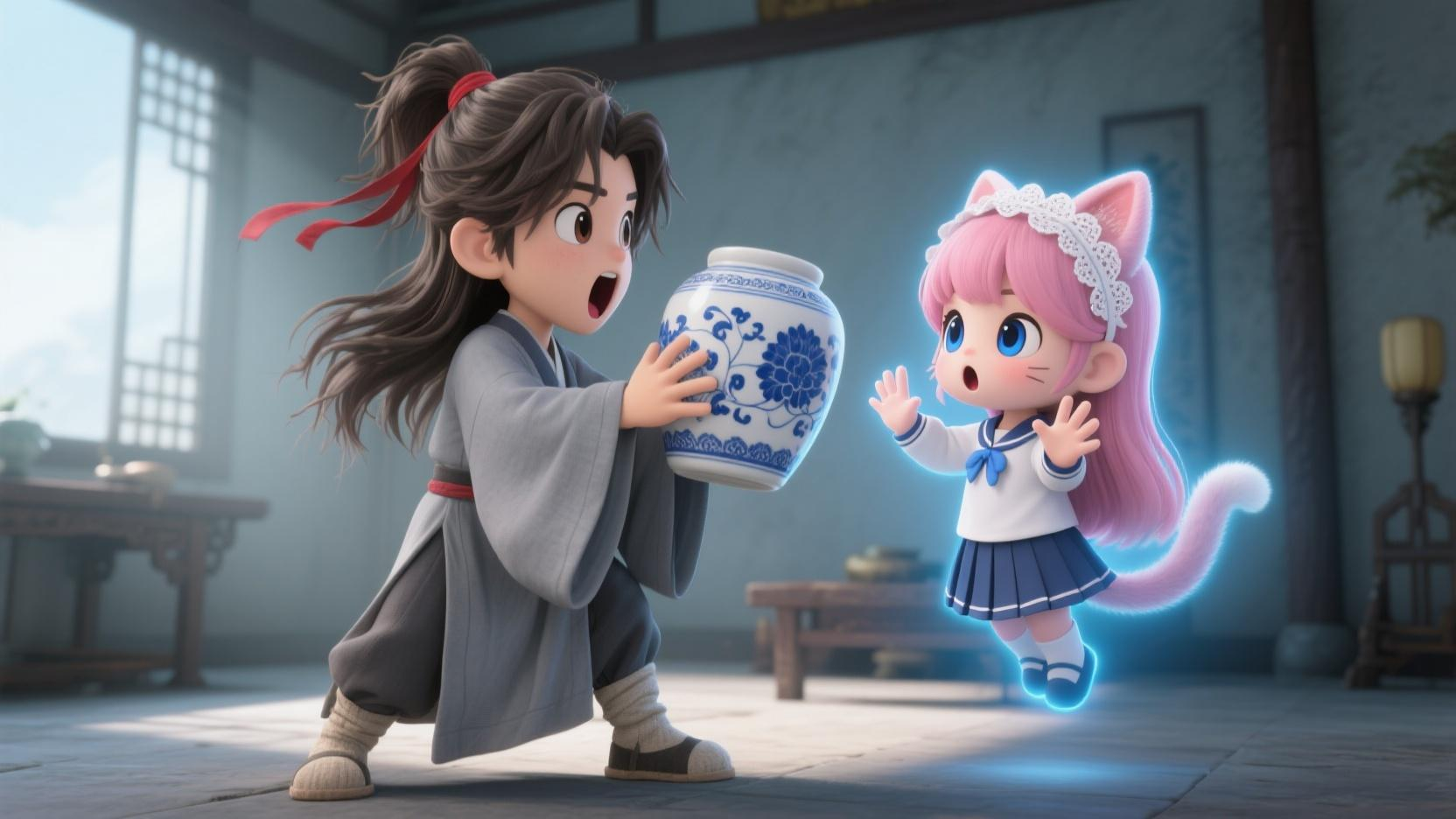}
        \caption{sceneYX01\_shot\_02}
        \label{fig:sceneYX01_shot_02}
    \end{subfigure}
    \hfill
    \begin{subfigure}[b]{0.32\textwidth}
        \includegraphics[width=\textwidth]{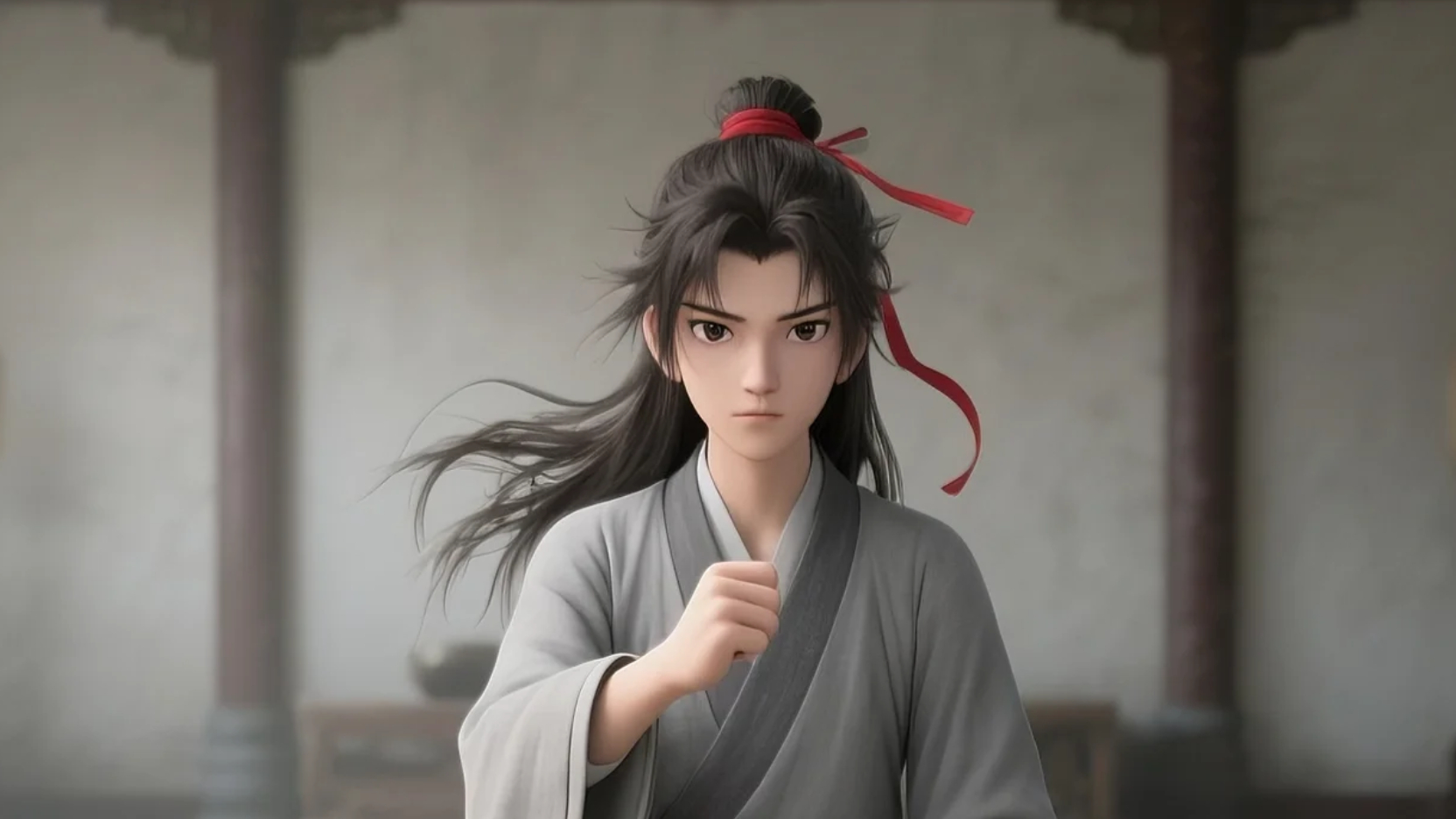}
        \caption{sceneYX01\_shot\_03}
        \label{fig:sceneYX01_shot_03}
    \end{subfigure}

    \caption{Storyboard keyframes for scene \texttt{sceneYX01}. Each shot shows the corresponding keyframe generated by the Storyboard Agent using selected tools (text-to-image, reference image, layout-guided generation).}
    \label{fig:sceneYX01_storyboard}
\end{figure*}

In AniME, the Model Context Protocol (MCP) enables specialized agents to autonomously select and invoke the most appropriate tools for their tasks based on the context provided by the Director. As an example, consider a scene description sent from the Director to the Storyboard Agent:

\begin{quote}
\emph{``In Ye’s training room, Ye raises a blue-and-white porcelain cup with both hands, tilting his head to bring it to his mouth, while the system AI angrily try to stop him, telling him to stop.''}
\end{quote}

Upon receiving this shot description, the Script \& Storyboard Agent first generate the following storyboard descriptions and choose the image generation tools.
For each shot, the Storyboard Agent produces a structured JSON output, including the selected tool, prompts, references, and notes. Example:

\begin{verbatim}
{
  "scene_id": "scene_YX01",
  "storyboard_shots": [
    {
      "shot_id": "scene_YX01_shot_01",
      "tool": "reference_image_generation",
      "prompt": "Ye holding a blue-and-white porcelain cup,
       tilting head to drink",
      "reference_images": ["assets/char_YX_front.png"],
      "notes": "Preserves character identity"
    },
    {
      "shot_id": "scene_YX01_shot_02",
      "tool": "layout_guided_generation",
      "prompt": "System AI angrily stopping Ye",
      "layout_bboxes": [
            {
            "object": "Ye Xuan",
            "bbox": [100, 300, 400, 900],
            "notes": "Left side of the frame; full body visible"
            },
            {
            "object": "System AI",
            "bbox": [600, 350, 900, 850],
            "notes": "Right side of the frame; mid-body
             visible, pointing gesture"
            }
        ],
      "notes": "Precise spatial relationship between characters"
    },
    {
      "shot_id": "scene_YX01_shot_03",
      "tool": "layout_guided_generation",
      "prompt": "Close-up of Ye's facial expression reacting to AI",
      "layout": "assets/layouts/scene_YX01_shot04.json",
      "notes": "Detailed facial expression and posture"
    }
  ]
}
\end{verbatim}

Through MCP, the Storyboard Agent adapts its tool selection based on the task requirements and scene context, ensuring efficient and high-quality storyboard generation without manual intervention. The Director’s role remains providing the high-level scene context, while MCP governs how the specialized agent internally orchestrates its tools.


\end{document}